\documentclass[preprint,12pt]{elsarticle}

\usepackage{amsmath,amsfonts}
\usepackage{algorithmic}
\usepackage{array}
\usepackage[caption=false,font=normalsize,labelfont=sf,textfont=sf]{subfig}
\usepackage{textcomp}
\usepackage{stfloats}
\usepackage{url}
\usepackage{verbatim}
\usepackage{graphicx}
\usepackage{balance}
\usepackage{bbm}
\usepackage{lipsum}
\usepackage{amssymb}
\usepackage{amsmath}
\usepackage{tabularx}
\usepackage{booktabs}
\usepackage{color, colortbl}
\usepackage{graphicx}
\usepackage{caption}
\usepackage{multirow}
\usepackage{subcaption}
\usepackage{xspace}
\usepackage{lineno}
\usepackage[dvipsnames]{xcolor}
\usepackage{makecell}
\usepackage{lineno}
\DeclareMathOperator*{\argmax}{arg\,max}

\usepackage[colorlinks=true, linkcolor=blue, citecolor=blue, urlcolor=blue]{hyperref}
\journal{Expert Systems with Applications}

\begin{document}

\begin{frontmatter}



\title{E-InMeMo: Enhanced Prompting\\ for Visual In-Context Learning}


\author[osu]{Jiahao Zhang}
\ead{jiahao@is.ids.osaka-u.ac.jp}

\author[osu]{Bowen Wang\corref{cor1}}
\ead{wang@ids.osaka-u.ac.jp}

\author[xmu]{Hong Liu}
\ead{lynnliu.xmu@gmail.com}

\author[osu,meetyou]{Liangzhi Li}
\ead{li@ids.osaka-u.ac.jp}

\author[sanken]{Yuta Nakashima}
\ead{n-yuta@ids.osaka-u.ac.jp}

\author[osu]{Hajime Nagahara}
\ead{nagahara@ids.osaka-u.ac.jp}

\address[osu]{D3 Center, The University of Osaka, Osaka, 565-0871, Japan}
\address[sanken]{SANKEN, The University of Osaka, Osaka, 567-0047, Japan}
\address[xmu]{School of Informatics, Xiamen University, Xiamen, 361000, China}
\address[meetyou]{Meetyou AI Lab, Xiamen Meet You Co., Ltd, Xiamen, 361000, China}

\cortext[cor1]{Corresponding author}

\begin{abstract}
Large-scale models trained on extensive datasets have become the standard due to their strong generalizability across diverse tasks. In-context learning (ICL), widely used in natural language processing, leverages these models by providing task-specific prompts without modifying their parameters. This paradigm is increasingly being adapted for computer vision, where models receive an input-output image pair, known as an in-context pair, alongside a query image to illustrate the desired output. However, the success of visual ICL largely hinges on the quality of these prompts. To address this, we propose \textbf{\underline{E}}nhanced \textbf{\underline{In}}struct \textbf{\underline{Me}} \textbf{\underline{Mo}}re (E-InMeMo), a novel approach that incorporates learnable perturbations into in-context pairs to optimize prompting. Through extensive experiments on standard vision tasks, E-InMeMo demonstrates superior performance over existing state-of-the-art methods. Notably, it improves mIoU scores by 7.99 for foreground segmentation and by 17.04 for single object detection when compared to the baseline without learnable prompts. These results highlight E-InMeMo as a lightweight yet effective strategy for enhancing visual ICL. Code is publicly available at: \url{https://github.com/Jackieam/E-InMeMo}
\end{abstract}




\end{frontmatter}


\section{Introduction}
Recent years have witnessed significant progress in large-scale models, which have shown impressive generalization capabilities and strong potential across various downstream tasks \cite{largemodel, dosovitskiy2020image, clip}. Representative models like GPT \cite{gpt4o} and Gemini \cite{gemini} exemplify the effectiveness of \textit{in-context learning} (ICL) in Natural Language Processing (NLP) \cite{rubin2021learning, gonen2022demystifying, wu2022self, sorensen2022information, honovich2022instruction, wang2022self, min2021noisy, vascar, dirct}. Through ICL, models can tackle novel tasks by leveraging prompt-based inputs to infer predictions on previously unseen data, without requiring updates to their parameters, thus substantially reducing training costs. Despite its promise as a foundational paradigm for deploying large-scale models in real-world scenarios, the exploration of ICL in computer vision applications is still at a relatively early stage \cite{maevqgan, supicl, wang2023images}.

\begin{figure}[t]
  \centering
   \includegraphics[width=0.7\linewidth]{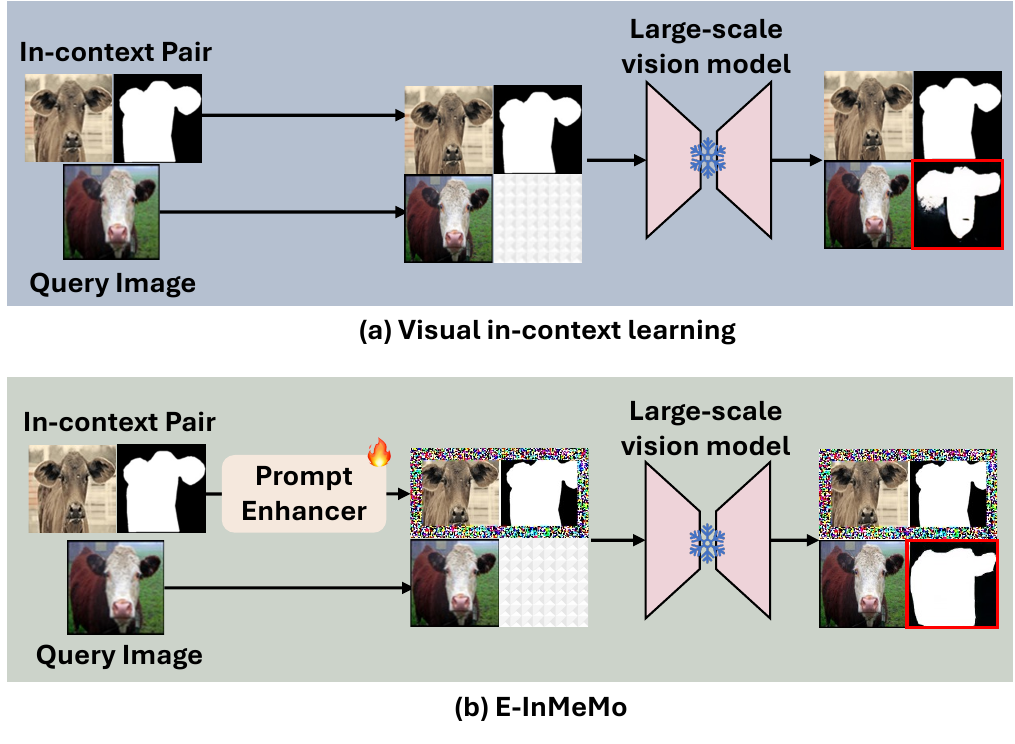}
   \caption{A schematic comparison between standard visual ICL and E-InMeMo. (a) \textcolor{RoyalBlue}{Visual ICL} constructs a four-cell grid canvas composed of a query image, an in-context pair, and an empty cell (bottom-right) where the model generates the prediction. This canvas serves as the \textit{prompt}, and the output (marked in the red box) is produced by passing it through a frozen large-scale vision model. (b) \textcolor{Orange}{E-InMeMo} enhances this paradigm by introducing a \textit{learnable prompt}, a trainable perturbation designed to adjust the distribution of the in-context pair, thereby improving task guidance and prediction accuracy.}
   \label{fig:icl_concept}
\end{figure}

\begin{figure}[t]
  \centering
   \includegraphics[width=0.7\linewidth]{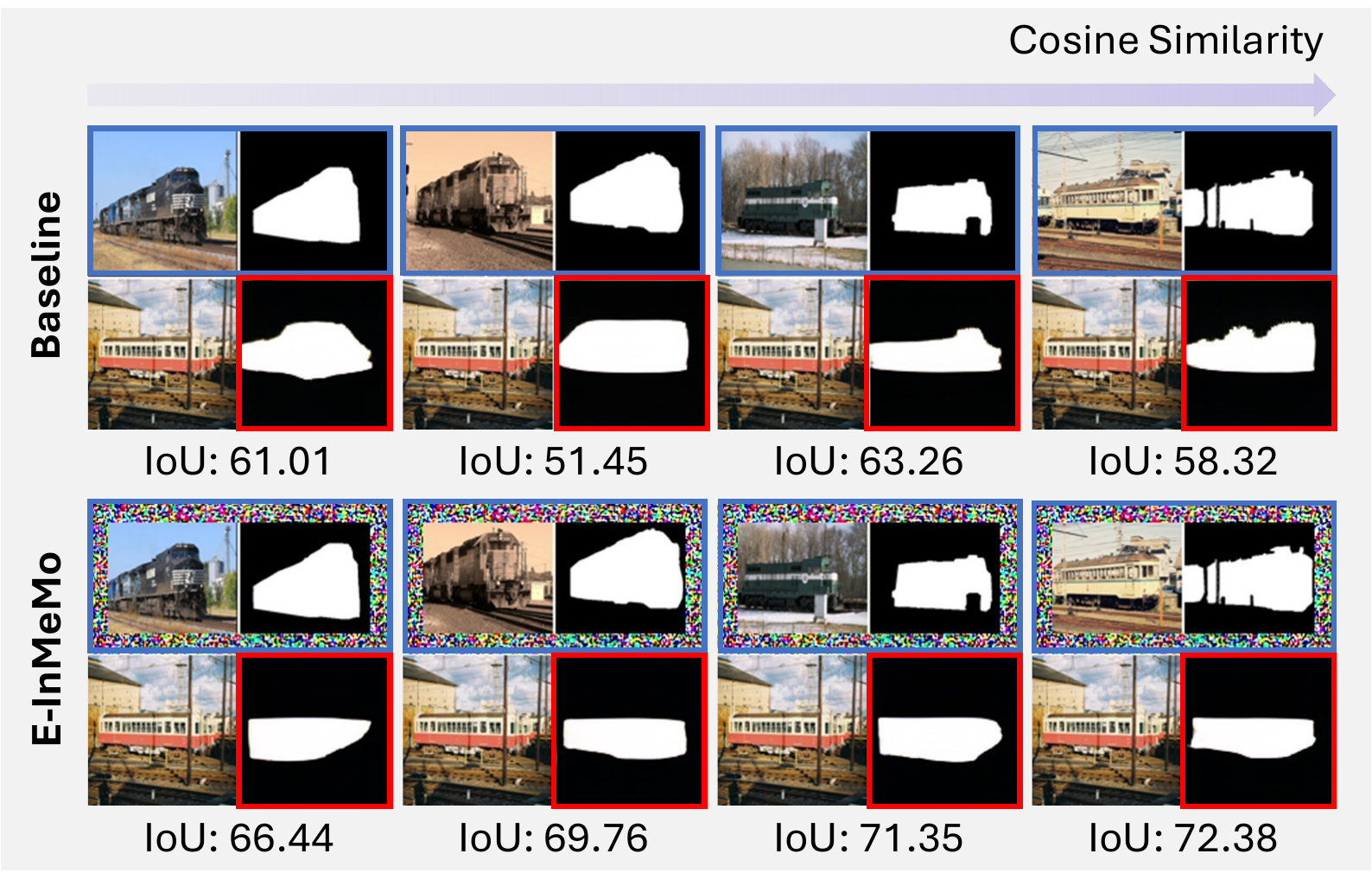}
   \caption{Performance comparison of visual ICL on a foreground segmentation task. \textcolor{RoyalBlue}{Blue boxes} indicate in-context pairs, while \textcolor{red}{red boxes} represent the predicted label images (query images are unmarked). The quality and similarity of the in-context pair significantly influence the prediction results. In the absence of a learnable prompt, performance is highly dependent on the semantic closeness between the query and in-context images. In contrast, E-InMeMo, with its learnable prompt, produces more stable and accurate predictions across varying input conditions.}
   \label{fig:prompt_selection}
\end{figure}

MAE-VQGAN \cite{maevqgan} marks a groundbreaking development by showcasing the potential of applying ICL to a range of computer vision tasks, such as image segmentation, inpainting, and style transfer. This method introduces visual prompts arranged in a four-cell grid canvas, as shown in Figure~\ref{fig:icl_concept}(a), which includes a query image alongside an input-output example known as an in-context pair. These in-context pairs illustrate the task at hand by pairing an input image with its corresponding label or output image. Prior work \cite{supicl} has emphasized the importance of such pairs in effectively guiding the model toward producing accurate results. In visual ICL, it is particularly important that the in-context image closely resembles the query image in terms of semantics, perspective, and other attributes \cite{supicl}, as illustrated in Figure~\ref{fig:prompt_selection}. Consequently, selecting appropriate in-context pairs becomes a critical component of the overall process.

Although visual ICL methods have shown promising results \cite{supicl, promptself}, the in-context pairs retrieved are not always optimal. This is often due to limitations in the available retrieval dataset and a mismatch between the prompt content and the latent knowledge embedded within large-scale vision models. This raises a critical question: Is it possible to refine or transform the prompt to better guide the model for downstream visual ICL tasks?

The concept of learnable prompting\footnote{In \cite{firstvp}, modifying images with learnable pixel-level changes is called visual prompting. Since our approach also uses visual prompts consisting of a query and an in-context example, we refer to these pixel-level modifications as a learnable prompt.} \cite{firstvp, ilmvp, blackvip} provides a promising solution. It adjusts the model's input without altering the model parameters, enabling adaptation to various tasks. This method, known as parameter-efficient fine-tuning (PEFT) \cite{vpt, lester2021power, prompttuning2, coop}, is especially suitable for large models where conventional fine-tuning can be prohibitively expensive due to their scale \cite{ilmvp, firstvp, elsayed2018adversarial, neekhara2022cross}. Learnable prompts have demonstrated a remarkable ability to adapt, even in cases with significant semantic or distributional differences \cite{blackvip, ilmvp, firstvp}.

In this work, we introduce a novel visual ICL method called \textbf{\underline{E}nhanced} \textbf{\underline{In}}struct \textbf{\underline{Me}} \textbf{\underline{Mo}}re (\textbf{E-InMeMo}), which leverages learnable visual prompts to instruct large vision models. After retrieving an initial in-context pair, we refine it using a dedicated prompt enhancer inspired by \cite{firstvp}. As with previous visual ICL approaches, the refined in-context pair is combined with the query into a unified image canvas, which is then processed by a pre-trained large-scale vision model \cite{maevqgan}. The learnable prompt is trained in a supervised fashion to produce the correct label image for the query.

\textbf{Contributions.} E-InMeMo offers a PEFT-based solution that enables efficient training while achieving high performance. By customizing a learnable prompt for a given task, the framework reshapes the prompt distribution to be more aligned with the task objectives, thereby improving both the encoding and decoding stages within the large-scale model. Our experiments establish new state-of-the-art (SOTA) results in tasks such as foreground segmentation and single object detection. Although E-InMeMo requires training, it substantially mitigates the limitations posed by suboptimal visual prompts.

This paper extends our earlier conference work published at WACV 2024 \cite{inmemo}. The enhancements in this version include: 1) introducing a new strategy to apply learnable prompt directly to the in-context pair, with strong robustness; 2) removing the need for resizing the in-context pairs, and thus reducing trainable parameters from 69,840 to 27,540; 3) replacing CLIP \cite{clip} with DINOv2 \cite{dinov2} for feature extraction, improving representation quality; and 4) validating E-InMeMo’s consistent outperformance across diverse datasets, including natural and medical images. Furthermore, we present comprehensive experiments to highlight E-InMeMo’s robustness and generalizability.

\section{Related Work}
\subsection{In-Context Learning}

In-Context Learning (ICL) has emerged as a powerful framework in Natural Language Processing (NLP), particularly with the advent of large language models (LLMs) such as GPT-3 \cite{gpt3}. By presenting several input-output examples pertaining to a specific task, this paradigm enables autoregressive models to perform inference without adjusting their parameters \cite{supicl}. ICL has gained significant traction due to its various strengths \cite{dong2022survey}, including an intuitive and interpretable user interface \cite{gpt3, liu2021makes, lu2021fantastically}, its resemblance to human-like reasoning \cite{winston1980learning}, and its suitability for plug-and-play deployment as a service \cite{sun2022black}. It has opened doors to new possibilities across different areas \cite{iclnlp1, learn2learn, kim2022self}, such as solving complex reasoning tasks \cite{mathicl}, question answering \cite{learn2learn, press2022measuring}, and compositional generalization \cite{an2023context, hosseini2022compositional}.

In the computer vision field, ICL has sparked growing interest \cite{alayrac2022flamingo, supicl, maevqgan, wang2023images, lvm}. Unlike NLP, where instructions are text-based, visual ICL presents a unique challenge: how to clearly define the task for the model using visual inputs. Bar \textit{et al.} \cite{maevqgan} introduced a novel method that combines an input-output image pair with a query image, to specify the task. This setup, presented as a single image, reframes the task as a form of image inpainting task. Subsequent efforts like Painter \cite{wang2023images}, which adopts the MAE architecture instead of VQGAN, and SegGPT \cite{seggpt}, which generalizes the method to image and video segmentation, have built upon this setting. LVM \cite{lvm} further pushes the boundaries by tokenizing visual inputs and applying an autoregressive learning strategy akin to that used in LLMs.

Beyond developing new architectures, recent studies underscore the critical role of in-context image quality in determining performance \cite{supicl, promptself}. Zhang \textit{et al.} \cite{supicl} proposed a supervised approach for prompt selection, emphasizing that both the number and the relevance of in-context pairs significantly affect results. Similarly, Sun \textit{et al.} \cite{promptself} explored pixel-level retrieval methods for selecting in-context pairs, experimenting with multiple layout configurations to improve ICL effectiveness.

While existing research has firmly established the importance of in-context pairs for downstream tasks \cite{maevqgan, supicl, promptself}, little attention has been paid to altering or enhancing these pairs to boost visual ICL performance. Our study seeks to fill this gap by introducing learnable perturbations to refine in-context pairs and ultimately enhance downstream outcomes.

\subsection{Learnable Prompting}

In the NLP field, prompting is widely used to adapt LLMs for specific downstream applications \cite{liu2023pre}. For example, while GPT-3 \cite{gpt3} demonstrates strong generalization across many tasks, it often relies on carefully engineered prompts, an approach that can be labor-intensive. Additionally, fully fine-tuning these massive models incurs high computational costs. Parameter-efficient fine-tuning (PEFT) addresses these challenges by modifying only a small fraction of the model parameters or by adding auxiliary components, such as adapters \cite{pfeiffer2020adapterfusion, houlsby2019parameter, zhang2023llama} or learned prompt vectors \cite{lester2021power, hu2021knowledgeable}, achieving performance comparable to full fine-tuning with far less overhead.

Motivated by the success of PEFT in NLP, researchers have extended these strategies to computer vision \cite{vpt, tsai2020transfer, blackvip, chen2022adaptformer} and vision-language models \cite{clip, coop, cocoop, gao2024clip}. A common approach involves augmenting the input image with learnable prompt embeddings or perturbations. Several recent methods have demonstrated how trainable modifications to visual inputs can enhance model performance \cite{firstvp, ilmvp, tsao2023autovp}. For instance, Bahng \textit{et al.} \cite{firstvp} proposed adding learnable, input-agnostic pixel-level prompts to input images to better adapt large, frozen models like CLIP \cite{clip} to new tasks. ILM-VP \cite{ilmvp} extended this by introducing task-aware prompts that incorporate label mapping for improved transfer learning. Since this method involves optimizing only a small number of parameters relative to the full model, it offers a practical and scalable solution—particularly well-suited for integration into visual ICL pipelines. In this work, we explore the integration of visual prompting to improve task performance in vision models using in-context learning.

\begin{figure*}[t]
  \centering
   \includegraphics[width=1\linewidth]{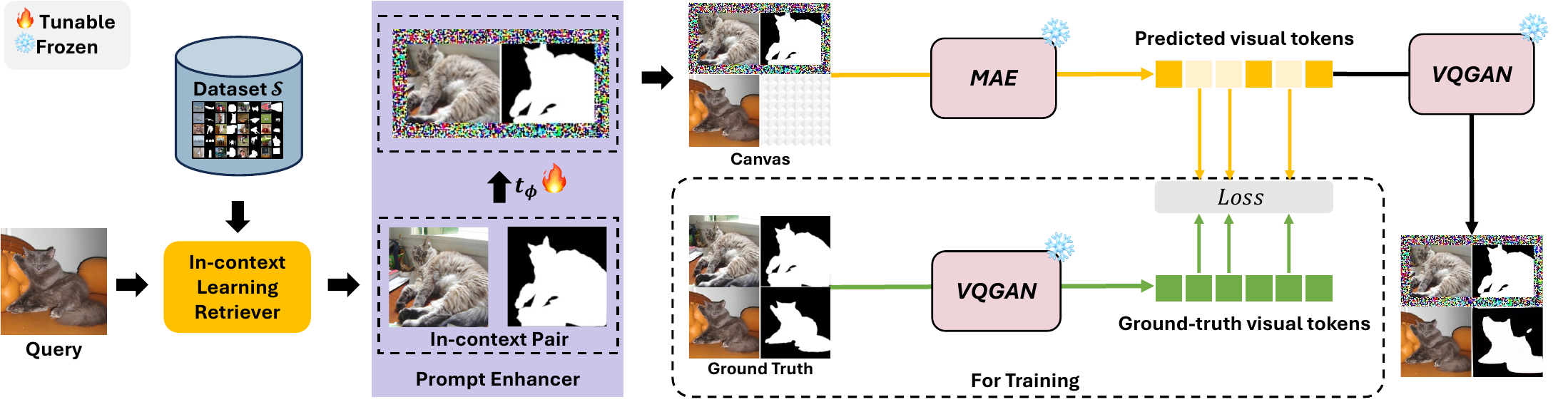}
   \caption{Overview of the proposed E-InMeMo framework. The process begins with the In-context Learning Retriever, which selects an in-context pair from the dataset $\mathcal{S}$ for a given query image. A \textcolor{Plum}{Prompt Enhancer $t_\phi(\cdot)$} then applies learnable perturbations to the in-context pair, producing an enhanced version. These enhanced in-context images, along with the query and an empty cell (bottom-right), are arranged into a four-cell grid canvas. This canvas is passed through a frozen {\tt MAE}, which outputs predicted visual tokens corresponding to the empty cell. For visualization, the predicted tokens are decoded into an image using the decoder of {\tt VQGAN}. During training, a ground-truth canvas, comprising the original in-context pair and the true label for the query, is encoded using a pre-trained {\tt VQGAN} encoder to produce ground-truth tokens. A cross-entropy loss is computed on the empty cell to update only the parameters of the Prompt Enhancer, $\phi$.}
   \label{fig:inmemo}
\end{figure*}

\section{Method}

\subsection{Preliminary: MAE-VQGAN}

MAE-VQGAN represents a pioneering effort that first introduces visual ICL via inpainting \cite{maevqgan}. Consider a query image denoted by $x_\text{q} \in \mathbb{R}^{C \times H \times W}$, alongside an example input $x \in \mathbb{R}^{C \times H \times W}$ and its corresponding ground-truth output $y \in \mathbb{R}^{C \times H \times W}$ (\textit{e.g.}, a segmentation result). These components form what is referred to as \textit{prompt}. The prompt is created by concatenating $x_\text{q}$, $x$, and $y$ into a single canvas $c_\text{q} = [x, y, x_\text{q}, r] \in \mathbb{R}^{C \times 2(H+1) \times 2(W+1)}$,\footnote{Following \cite{maevqgan}, we add a two-pixel gap between images.} where $[\cdot, \cdot, \cdot, \cdot]$ represents image concatenation to formulate a four-cell grid canvas. The region $r \in \mathbb{R}^{C\times H \times W}$ remains blank, designated for the task output $\hat{y}_\text{q}$.\footnote{In \cite{maevqgan}, the arrangement of $x_\text{q}$, $x$, and $y$ is flexible. A mask is used to specify the region $r$ to be filled. For brevity, these details are omitted here.} The region $r$ is subdivided into $L$ patch regions, forming a patch set $\mathcal{L} = \{r_l\}_{l=1}^L$. The MAE-VQGAN computes the \textit{assignment scores} vector for the patch region $r_l$ as follows:
\begin{align}
    s_{l} = g_{l}(c_\text{q}) \quad \in \mathbb{R}^{|\mathcal{V}|},
\end{align}
where $\mathcal{V}$ represents the pre-trained VQGAN codebook, and $g_{l}$ is the MAE function that outputs the assignment score of $r_l$, with each element $s_{lv} \in s_{l}$ indicating the score for visual token $v \in \mathcal{V}$. These scores can be interpreted as the probability $p(w_l = v|c_\text{q})$ that the visual token assigned to $l$, denoted by $w_l$, is $v$. After determining the appropriate visual token for $l$ as $z^\star_l = \argmax_v s_{lv}$, the VQGAN decoder $f$ generates the prediction $\hat{y}_q$, expressed as:
\begin{align}
    \hat{y}_q = f(\{z^\star_l\}_{l=1}^L) \quad \in \mathbb{R}^{C\times H \times W}.
\end{align}

\subsection{Overview of E-InMeMo}

Let $\mathcal{S} = {(x, y)}$ represent a dataset consisting of input images $x$ paired with their corresponding label (or output) images $y$, designated for a specific downstream task, where $|\mathcal{S}| = n$. Given this dataset and a query image $x_\text{q}$, the goal is to predict the output $y_\text{q}$ for the query.

An outline of the proposed E-InMeMo framework is illustrated in Figure~\ref{fig:inmemo}. First, the query image $x_\text{q}$ is passed through an retriever, which selects a similar in-context pair $(x, y)$ from the dataset $\mathcal{S}$. This selected pair is combined with the query image into a $2\times2$ grid format, forming what we refer to as a canvas. The canvas is expressed as $c_\text{q} = [x, y, x_\text{q}, r]$, where $r$ denotes an empty cell.

The prompt enhancer, which integrates learnable prompt, then refines the in-context pair to produce $(x', y')$. This results in an enhanced version of the canvas: $\hat{c}_\text{q} = [x', y', x_\text{q}, r]$. This enhanced canvas is then input to a frozen, pre-trained large-scale vision model $g$, which outputs a sequence of visual tokens $\hat{z} = g(\hat{c}_\text{q})$. Simultaneously, the complete ground-truth canvas $c_\text{gt} = [x, y, x_\text{q}, y_\text{q}]$ is encoded using a VQGAN encoder to generate the reference visual tokens $z = g(c_\text{gt})$.

To optimize the learnable prompt, we compute a cross-entropy loss over the visual tokens corresponding to the empty cell location. These tokens encode the model’s prediction $\hat{y}_\text{q}$, which is decoded by a function $f$ that maps tokens back into the pixel domain, producing the final output $\hat{y}_\text{q} = f(\hat{z})$.

At the heart of E-InMeMo lies the prompt enhancer, denoted as $t_\phi$, parameterized by a learnable set $\phi$. This component is trained on dataset $\mathcal{S}$ to generate instructive and task-specific transformations, ensuring effective guidance even in scenarios where the retrieved in-context pair is suboptimal. 

\subsection{In-context Learning Retriever}

Selecting a high-quality in-context pair that aligns well with a given query image is essential for enhancing ICL performance \cite{supicl}. To improve this selection process, we advocate for leveraging rich and comprehensive visual features during retrieval. To this end, we propose a Feature-Map-Level Retrieval (FMLR) strategy.

In particular, we employ a pre-trained visual encoder, such as the one from DINOv2 \cite{dinov2} to extract $\ell_2$-normalized visual features for both the query image $x_\text{q}$ and every candidate image $x \in \mathcal{S}$. Unlike methods that rely on global features, FMLR preserves the full feature map, flattens it into a 1-dimensional representation, and utilizes it directly for similarity comparison.

Among all available in-context candidates in $\mathcal{S}$, the one whose visual representation is most similar to that of the query is selected as the in-context pair $(x, y)$. Formally, the selection is given by:
\begin{equation}
  (x, y) = \argmax_{(x^\star, y^\star) \in S} e(x_\text{q})^\top e(x^\star),
  \label{eq:prompt_ret}
\end{equation}
where $e(\cdot)$ denotes the extracted and normalized visual features.

\subsection{Prompt Enhancer}

Learnable visual prompting, initially introduced in \cite{firstvp}, draws inspiration from prompt-based learning in NLP \cite{gpt3, liu2023pre, han2022ptr}. It offers an effective solution for addressing domain shifts by enabling input adaptation for downstream tasks, without requiring any fine-tuning of the base model. In our setting, we follow the formulation in \cite{firstvp}, introducing pixel-level learnable perturbations near the image boundaries to guide the model more effectively.

In the E-InMeMo framework, the Prompt Enhancer applies such learnable prompts to the in-context image pair. These enhanced examples provide additional instructions that help the frozen model interpret the task more accurately, thereby narrowing the domain gap between the query image and the provided in-context pairs. Notably, the learnable prompt is input-agnostic; a single shared prompt is used across all in-context pairs for the same task. This allows the learned prompt to act as a task-specific identifier.

Given a retrieved in-context pair $(x, y)$, the prompt enhancer applies a learnable perturbation $t_{\phi}$, controlled by a parameter set $\phi$, to produce the modified pair $(x', y')$ as follows:
\begin{equation}
  (x', y') = (x, y) + \delta t_\phi,
  \label{eq:vp_on_sp}
\end{equation}
where $\delta$ defines the perturbation's strength. The prompt $t_\phi$ resides in the image space and consists of learnable pixels concentrated along the image borders. All other pixels are zeroed out. These parameters are optimized through backpropagation during training.

\subsection{Prediction}

Following \cite{maevqgan}, we adopt the MAE-VQGAN model, in which pre-trained MAE \cite{mae} $g$ generates visual tokens $\hat{z}$ from $\hat{c}_\text{q}$. The VQGAN \cite{vqgan} decoder $f$, again pre-trained, generates resulting image $\hat{y}_\text{q}$ from $\hat{z}$. 
 
After compiling the in-context pair and the query into a canvas $\hat{c}_\text{q}$, $g$ predicts latent visual tokens $\hat{z} = (\hat{z}_1,\dots,\hat{z}_L)$, specifically,
\begin{align}
    \hat{z}_l = \argmax_v g_{lv}(\hat{c}_\text{q}), \label{eq:mae_encoding}
\end{align}
where $\hat{z}_l \in \hat{z}$ is a visual token in the vocabulary $\mathcal{V}$ at spatial position $l$, and $g_{lv}$ gives the probability of $v \in \mathcal{V}$ for $l$. The VQGAN decoder $f$ then generates a label image by 
\begin{align}
    \hat{y}_\text{q} = f(\hat{z}).
\end{align} 
We obtain the prediction for the query $x_\text{q}$ as $\hat{y}_\text{q}$. 

\subsection{Training}

The only learnable parameters in E-InMeMo are the prompt $t_\phi$. We train it for a specific task on $\mathcal{S}$. The loss is the same as \cite{maevqgan}, while all parameters except for $t_\phi$ are frozen. 

We first randomly choose a pair $(x_\text{q}, y_\text{q})$ as query from $\mathcal{S}$. The E-InMeMo prediction process from the ICL retriever is then applied to $x_\text{q}$ to compute $\hat{z}$, but the retriever uses $\mathcal{S}\setminus\{(x_\text{q}, y_\text{q})\}$ instead of $\mathcal{S}$. 

The label image $y_\text{q}$ is used for training. We compile the retrieved in-context pair $(x, y)$ and $(x_\text{q}, y_\text{q})$ into a canvas $c_\text{gt} = (x, y, x_\text{q}, y_\text{q})$.
The pre-trained VQGAN encoder $E$ associated $f$ gives the ground-truth visual tokens $z$ that reconstruct $y_\text{q}$ with $f$, \textit{i.e.},
\begin{align}
    z_l = \argmax_v E_{lv}(c_\text{gt}),
\end{align}
where $E_{lv}$ again is the probability of $v \in \mathcal{V}$ for position $l$.
The loss function $\mathcal{F}$ to train our learnable prompt $t_\phi$ is given by
\begin{align}
    \mathcal{F}(\phi) = \mathbb{E}[\text{CE}(g_{l}(\hat{c}_\text{q}), z_l)], \label{eq:loss}
\end{align}
where $\text{CE}$ is the cross-entropy loss, $g_{l}(\hat{c}_\text{q}) \in \mathbb{R}^{|\mathcal{V}|}$ is the probabilities of respective tokens in $\mathcal{V}$, and the expectation is computed over all $(x_\text{q}, y_\text{q}) \in \mathcal{S}$ as well as all visual tokens $z_l$ corresponding to $y_\text{q}$ (\textit{i.e.} over the latent visual tokens of $r$, represented as masked index).

\subsection{Interpretation}

Adding $t_\phi$ to images in a visual prompt as in Equation~(\ref{eq:vp_on_sp}) translates the distribution of the prompt in a certain direction. Determining $t_\phi$ by Equation~(\ref{eq:loss}) will encode some ideas about the task described by $\mathcal{S}$ in $\phi$, supplying complementary information that is not fully conveyed by the in-context pair $(x, y)$. We consider that our training roughly aligns the distributions of image patches $\hat{c}_\text{q}$ and $c_\text{gt}$ in the latent space before visual token classification with smaller degrees of freedom in $t_\phi$. This can be particularly effective as these distributions are inherently different due to the lack of the ground-truth label image $y_\text{q}$ in the canvas. Therefore, our best expectation is that $\phi$ captures the distribution of $y_\text{q}$ collectively to bring the distribution of prompts closer to the ground-truth prompts (containing ground-truth label $y_\text{q}$). With this, the encoder $f$ will have better access to more plausible visual tokens that decode a label image closer to the ground-truth label.

\section{Experiments}

\subsection{Experimental Setup}

\textbf{Datasets and Downstream Tasks.}
We follow the experimental settings of \cite{maevqgan} to evaluate E-InMeMo. As downstream tasks, we perform foreground segmentation and single object detection. (1) \textbf{Foreground segmentation} aims to extract apparent objects from the query image with the in-context pair. We use the Pascal-5$^i$ dataset \cite{pascal}, which is split into four-fold subsets, each containing five classes. (2) \textbf{Single object detection} evaluates whether a model can capture fine-grained features specified by a coarse-grained bounding box in the in-context pair. We conduct experiments on images and bounding boxes from the PASCAL VOC 2012 \cite{everingham2010pascal}. To align with \cite{maevqgan}, we use a subset of the dataset whose samples only contain a single object as our dataset $\mathcal{S}$.

We also verify E-InMeMo on two medical datasets for foreground segmentation. (1) \textbf{Kvasir-SEG dataset} \cite{kvasir} is a dataset for gastrointestinal polyp segmentation. It contains 1000 polyp images and their corresponding binary segmentation masks. Since there has no official split of the test set, we randomly split the whole dataset in a five-fold, to conduct the experiments. (2) \textbf{Skin lesion dataset} \cite{isic} is a dermoscopic image benchmark challenge for diagnosis of skin cancer. It contains 900 images for training and 379 images for test. Each image contains their binary segmentation masks labeled by experts.

\textbf{Methods for Comparison.}  
All experiments utilize MAE-VQGAN \cite{maevqgan} as the pre-trained large-scale vision model. E-InMeMo is compared against SOTA methods for visual ICL. Below are the details of the comparison methods.
\begin{itemize}
\item \textbf{Random} \cite{maevqgan} means randomly retrieves in-context pairs from the training set. 
\item \textbf{UnsupPR} \cite{supicl} uses CLIP as the feature extractor to retrieve in-context pairs based on features after the classification header.
\item \textbf{SupPR} \cite{supicl} introduces contrastive learning to develop a similarity metric for in-context pair selection.
\item \textbf{SCS} \cite{scs} proposes a stepwise context search method to adaptively search the well-matched in-context pairs based on a small yet rich candidate pool.
\item \textbf{Partial2Global} \cite{xu2024towards} proposes a transformer-based list-wise ranker and ranking aggregator to approximately identify the global optimal in-context pair.
\item \textbf{Prompt-SelF} \cite{promptself} applies an ensemble of eight different prompt arrangements, combined using a pre-defined threshold of voting strategy.
\item \textbf{FMLR (CLIP)} means to utilizes CLIP as the feature extractor for feature-map-level retrieval.
\item \textbf{FMLR (DINOv2)} utilizes DINOv2 as the feature extractor for feature-map-level retrieval. 
\end{itemize}
Additionally, we compare E-InMeMo with few-shot segmentation methods derived from meta-learning, such as \textit{OSLSM} \cite{pascal} and \textit{co-FCN} \cite{cofcn}. For medical datasets, E-InMeMo is compared with visual ICL methods and FMLR using different feature extractors.

\begin{table*}[t]
    \centering
    \footnotesize
    \begin{tabularx}{\textwidth}{l>{\centering\arraybackslash}X>{\centering\arraybackslash}X>{\centering\arraybackslash}X>{\centering\arraybackslash}X>{\centering\arraybackslash}X>{\centering\arraybackslash}X}
        \toprule
        & ~ & \multicolumn{4}{c}{\textbf{Seg.} (mIoU)} & \textbf{Det.} \\
        \cmidrule(lr){2-6} 
        & Fold-0 & Fold-1 & Fold-2 & Fold-3 & Mean & (mIoU)  \\ 
        \midrule
        \textit{Meta-learning} \\
        OSLSM \cite{pascal} & 33.60 & 55.30 & 40.90 & 33.50 & 40.80 & -  \\ 
        co-FCN \cite{cofcn} & 36.70 & 50.60 & 44.90 & 32.40 & 41.10 & -  \\ 
        \midrule
        \textit{In-context learning} \\
        Random \cite{maevqgan} & 28.66 & 30.21 & 27.81 & 23.55 & 27.56 & 25.45  \\ 
        UnsupPR \cite{supicl} & 34.75 & 35.92 & 32.41 & 31.16 & 33.56 & 26.84  \\
        SupPR \cite{supicl} & 37.08 & 38.43 & 34.40 & 32.32 & 35.56 & 28.22  \\
        SCS \cite{scs} & - & - & - & - & 35.00 & -  \\
        Partial2Global \cite{xu2024towards} & 38.81 & 41.54 & 37.25 & 36.01 & 38.40 & 30.66  \\
        prompt-SelF \cite{promptself} & 42.48 & 43.34 & 39.76 & 38.50 & 41.02 & 29.83  \\
        FMLR (CLIP) & 35.69 & 37.24 & 33.98 & 33.13 & 35.01 & 27.27  \\
        FMLR (DINOv2) & 36.48 & 37.41 & 35.03 & 34.77 & 35.92 & 27.18  \\
        InMeMo \cite{inmemo} & 41.65 & \textbf{47.68} & 42.43 & 40.80 & 43.14 & 43.21  \\
        \cellcolor{RoyalBlue!10}E-InMeMo (Ours) & \cellcolor{RoyalBlue!10} \textbf{42.82} & \cellcolor{RoyalBlue!10} 46.97 & \cellcolor{RoyalBlue!10} \textbf{43.00} & \cellcolor{RoyalBlue!10} \textbf{42.83} & \cellcolor{RoyalBlue!10} \textbf{43.91} & \cellcolor{RoyalBlue!10} \textbf{44.22}  \\
        \bottomrule
    \end{tabularx}
    \caption{Performance of the foreground segmentation and single object detection downstream tasks. The best scores in \textit{in-context learning} are highlighted in \textbf{bold}. \textbf{Seg.}~and \textbf{Det.}~stand for the foreground segmentation and single object detection tasks, respectively.}
    \label{tab:main results}
\end{table*}

\textbf{Implementation details.}
For the foreground segmentation task, E-InMeMo is trained separately on each fold of the training dataset, treating each fold as an independent task and learning a unique prompt for each one. In the case of single object detection, the model is trained using the entire training set, with in-context pairs being retrieved from the same set. During evaluation, each test image is treated as a query, and its corresponding in-context pair is selected from the training set.

All images are resized to 111 $\times$ 111 pixels to construct the canvas. The learnable prompting is applied around the borders of the in-context pair, occupying 15 pixels along each edge. This configuration results in a total of 27,540 trainable parameters for the prompt $\phi$. The canvas layout follows the standard format from \cite{maevqgan}, where the enhanced in-context pair $(x', y')$, the query image $x_\text{q}$, and an empty cell $r$ are placed at the top-left, top-right, bottom-left, and bottom-right positions, respectively. The perturbation strength $\delta$ is set to 1.

Our implementation is built with PyTorch. Training is conducted for 70 epochs using the Adam optimizer \cite{adam}, starting with an initial learning rate of 15. The learning rate follows a cosine annealing schedule with warm restarts. A key strength of E-InMeMo lies in its efficiency—the model can be trained on a single NVIDIA Tesla V100-SXM2-32GB GPU using a batch size of 32.

\subsection{Comparison with State-of-the-Art}

We present a comparative analysis of E-InMeMo against existing visual in-context learning (ICL) methods and meta-learning-based few-shot learning approaches, as shown in Table~\ref{tab:main results}. The results demonstrate that E-InMeMo sets a new SOTA in both downstream tasks, with particularly strong gains in single object detection. Notably, it significantly outperforms the FMLR baseline and consistently surpasses meta-learning-based methods across nearly all folds as well as in average performance. These findings underscore the effectiveness of incorporating a learnable prompt into the visual ICL framework.

One point of comparison is prompt-SelF \cite{promptself}, whose performance benefits from a bagging mechanism. It applies the model across eight different canvas configurations and aggregates the results, thereby leveraging diverse perspectives from the large vision model. In contrast, E-InMeMo requires only a single inference per query, yet achieves superior results. While bagging may offer a performance boost without significant overhead, the improvements achieved by E-InMeMo remain substantial.

E-InMeMo particularly excels in single object detection, where it exceeds the prompt-SelF by an impressive 13.38 points, highlighting its ability to accurately capture fine-grained visual cues and detect small objects with high precision.

These findings illustrate the strength of our direct and efficient strategy. By refining the in-context pair through learnable prompting, E-InMeMo effectively enhances the model's task comprehension. Furthermore, the approach remains lightweight, adding only 27,540 trainable parameters and requiring minimal computational resources. The task-specific, shared learnable prompts not only simplify the training process but also point toward a scalable and efficient direction for future visual ICL research. 

\begin{table*}[t]
    \centering
    \footnotesize
    {
    \begin{tabularx}{\textwidth}{l>{\centering\arraybackslash}X>{\centering\arraybackslash}X>{\centering\arraybackslash}X>{\centering\arraybackslash}X>{\centering\arraybackslash}X>{\centering\arraybackslash}X>{\centering\arraybackslash}X}
        \toprule
        ~ & \multicolumn{6}{c}{\textbf{Kvasir} (mIoU)} & \textbf{ISIC} \\
        \cmidrule(lr){2-7}
        \textbf{Method} & Fold-0 & Fold-1 & Fold-2 & Fold-3 & Fold-4 & Mean
        & (mIoU) \\
        \midrule
        Random \cite{maevqgan} & 37.79 & 33.98 & 35.72 & 32.33 & 32.51 & 34.47 & 62.43 \\ 
        UnSupPR \cite{supicl} & 39.71 & 39.99 & 41.02 & 35.90 & 37.54 & 38.83 & 68.78 \\ 
        SupPR \cite{supicl} & 41.16 & 40.94 & 41.58 & 38.53 & 36.33 & 39.71 & 68.76 \\
        FMLR (CLIP) & 41.17 & 40.91 & 40.49 & 39.40 & 37.01 & 39.80 & 70.77 \\
        FMLR (DINOv2) & 45.70 & 42.25 & 44.85 & 43.25 & 39.94 & 43.20 & 70.12 \\
        \rowcolor{RoyalBlue!10} E-InMeMo & \textbf{53.04} & \textbf{50.04} & \textbf{53.96} & \textbf{54.47} & \textbf{45.97} & \textbf{51.50} & \textbf{76.13} \\
        \bottomrule
    \end{tabularx}
    }
    \caption{The performance of E-InMeMo on ISIC and Kvasir datasets with different diseases.}
    \label{tab:med vp}
\end{table*}

\subsection{Results on Medical Datasets}
In addition to the dataset comprising natural images, we also verify E-InMeMo on two medical datasets as foreground segmentation task. The results are shown in Table~\ref{tab:med vp}. We observe that E-InMeMo also achieved the SOTA results on these two medical datasets, demonstrating E-InMeMo is also effective on medical domain dataset. We found that on ISIC dataset, there has a minor degradation of SupPR and FMLR (DINOv2) compared to suboptimal methods (such as UnSupPR and FMLR (CLIP). This phenomenon demonstrates previous methods cannot be ensured to get better results on different datasets (such as medical datasets). But E-InMeMo also showed optimal performance on such dataset transformation.  

\begin{table*}[t]
    \centering
    \footnotesize
    \resizebox{\textwidth}{!}{%
    \begin{tabular}{>{\centering\arraybackslash}m{3.2cm} >{\centering\arraybackslash}m{3.5cm}
                    >{\centering\arraybackslash}m{1cm} >{\centering\arraybackslash}m{1cm}
                    >{\centering\arraybackslash}m{1cm} >{\centering\arraybackslash}m{1cm}
                    >{\centering\arraybackslash}m{1cm}}
        \toprule
        Domain Setting & Method & Fold-0 & Fold-1 & Fold-2 & Fold-3 & Mean \\
        \midrule
        \multirow{2}{*}{\textit{Pascal $\rightarrow$ Pascal}} 
        & FMLR (DINOv2) & 36.48 & 37.41 & 35.03 & 34.77 & 35.92 \\
        & E-InMeMo
        & \cellcolor{RoyalBlue!10}42.82 
        & \cellcolor{RoyalBlue!10}46.97 
        & \cellcolor{RoyalBlue!10}43.00 
        & \cellcolor{RoyalBlue!10}42.83 
        & \cellcolor{RoyalBlue!10}\textbf{43.91} \\
        \midrule
        \multirow{2}{*}{\textit{COCO $\rightarrow$ Pascal}} 
        & FMLR (DINOv2) & 34.43 & 34.06 & 32.60 & 33.31 & 33.60 \\
        & E-InMeMo
        & \cellcolor{RoyalBlue!10}41.28 
        & \cellcolor{RoyalBlue!10}42.26 
        & \cellcolor{RoyalBlue!10}40.97 
        & \cellcolor{RoyalBlue!10}41.18 
        & \cellcolor{RoyalBlue!10}\textbf{41.42} \\
        \bottomrule
    \end{tabular}
    }
    \caption{Domain shifts analysis on E-InMeMo. \textit{Pascal $\rightarrow$ Pascal} means in-context pairs and query images are both from PASCAL. \textit{COCO $\rightarrow$ Pascal} indicates that in-context pairs are from COCO while query images are from PASCAL.}
    \label{tab:ds}
\end{table*}

\subsection{Domain Shifts Analysis}
In practical applications, domain shifts frequently occur, representing variations between the training dataset $\mathcal{S}$ and the target environment. These discrepancies can lead to reduced performance when models are applied beyond their original training distribution. These shifts are prevalent across datasets and are commonly used as benchmarks to evaluate the robustness of machine learning models \cite{domaing}.

To investigate E-InMeMo’s resilience to domain shifts, we adopt the COCO dataset \cite{coco} for out-of-domain inference, following the evaluation protocol established in \cite{supicl}. Specifically, we use a variant known as COCO-5$^i$, which aligns its category divisions with those of Pascal-5$^i$ \cite{supicl}. In this setup, referred to as COCO $\rightarrow$ Pascal, the in-context pairs are sampled from COCO-5$^i$, while the query images come from the Pascal-5$^i$ validation split, consistent with the configuration in \cite{promptself}.

The domain shifts evaluation results are summarized in Table~\ref{tab:ds}. Under the COCO $\rightarrow$ Pascal setting, the FMLR baseline using DINOv2 experiences a 2.32-point drop in mIoU, equivalent to a 6.46\% decrease. E-InMeMo, on the other hand, achieves an mIoU of 41.42\%, reflecting a 2.49-point drop (5.67\% decrease). Despite the slightly larger absolute drop, E-InMeMo exhibits a smaller relative decrease, and the performance gap between the two methods under domain shifts is minimal (0.17 points). This indicates that E-InMeMo maintains greater robustness in the face of domain shifts.

These findings suggest that incorporating learnable prompt into visual ICL not only improves in-domain performance but also enhances cross-domain generalization. The robustness and adaptability of E-InMeMo make it a promising approach for real-world applications where domain shifts are inevitable.

\subsection{Further Analyses on E-InMeMo}

This section further investigates the capabilities of E-InMeMo through a series of experiments, primarily focusing on the fine-grained foreground segmentation task. 

\begin{figure*}[!t]
  \centering
   \includegraphics[width=1\linewidth]{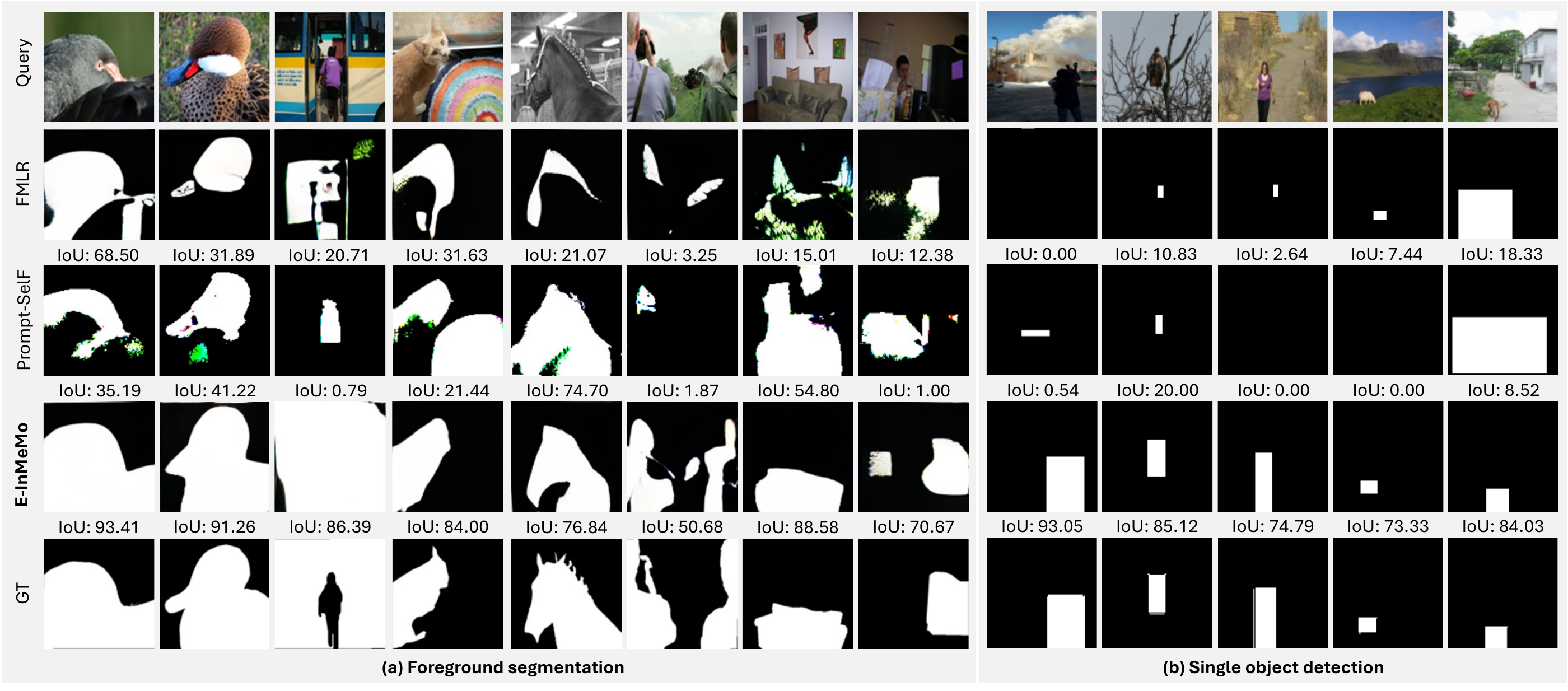}
   \caption{Qualitative comparisons across baseline methods, prompt-SelF, and our proposed E-InMeMo on two downstream tasks: \textbf{(a)} Foreground segmentation and \textbf{(b)} Single-object detection. For each task, the top row shows the query image. The subsequent rows (from top to bottom) present results from FMLR (DINOv2), prompt-SelF, \textbf{E-InMeMo}, and the ground-truth label (GT), respectively. E-InMeMo consistently enables visual ICL to capture finer details and demonstrates robustness to mismatches between in-context and query images. Notably, it also appears to mitigate the negative impact of low-quality in-context pairs—an important advantage when the retriever fails to find highly similar samples.}
   \label{fig:visual_examples}
\end{figure*}

\begin{figure*}[!t]
  \centering
   \includegraphics[width=1\linewidth]{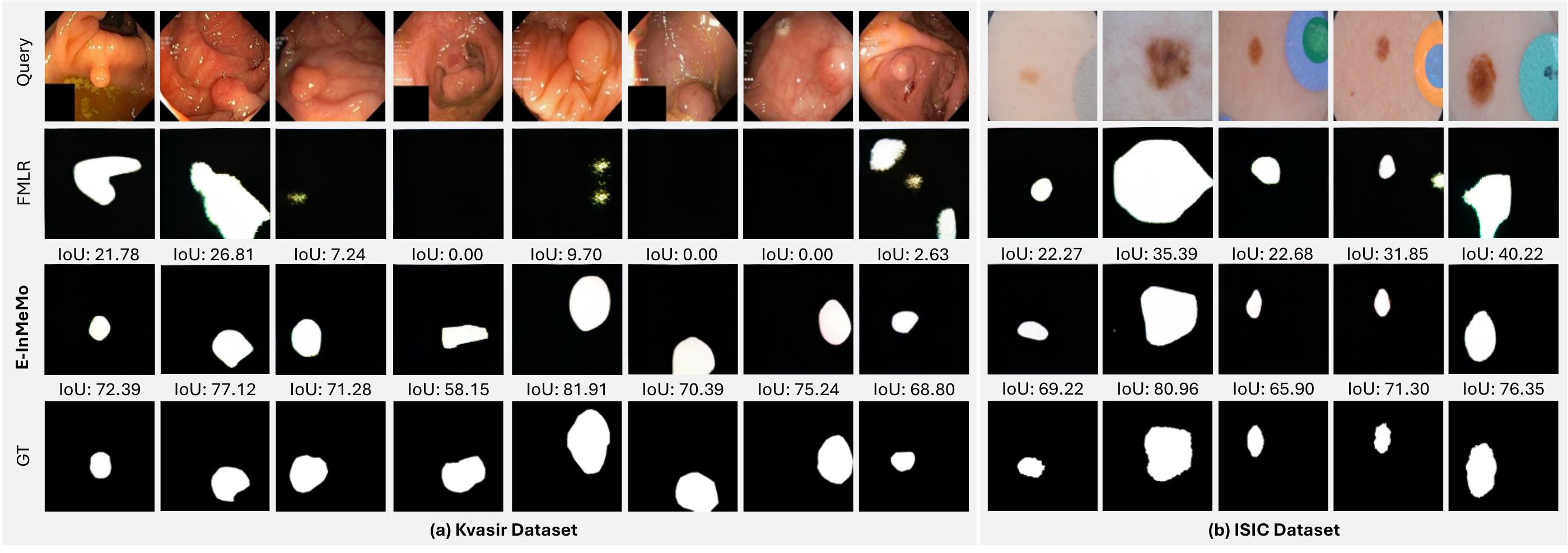}
   \caption{Qualitative results of FMLR (DINOv2) and our proposed E-InMeMo on two medical datasets: \textbf{(a)} Kvasir and \textbf{(b)} ISIC. For each dataset, the top row shows the query image. The following rows are arranged from top to bottom in the order of FMLR (DINOv2), \textbf{E-InMeMo}, and the ground-truth label (GT), respectively.}
   \label{fig:visual_examples_med}
\end{figure*}

\textbf{Qualitative comparison.} We conduct a qualitative comparison of E-InMeMo against FMLR (DINOv2), prompt-SelF \cite{promptself}, and the ground-truth (GT) labels across both foreground segmentation and single object detection tasks, as illustrated in Figure~\ref{fig:visual_examples}.

In the foreground segmentation task (Figure~\ref{fig:visual_examples}(a)), E-InMeMo produces segmentation results that closely align with the GT labels, capturing fine-grained details with high fidelity. Notably, E-InMeMo demonstrates strong robustness to various visual discrepancies. For instance, it performs well even when the in-context image is achromatic or when there is a substantial color difference between the in-context and query images. Additionally, E-InMeMo remains stable in cases involving significant foreground size variation and is capable of accurately distinguishing foreground from background in challenging scenes.

For the single object detection task (Figure~\ref{fig:visual_examples}(b)), E-InMeMo continues to exhibit consistent attention to detail. Its performance remains unaffected by changes in color or object scale within the in-context pair. Impressively, it is able to detect and localize objects even when the corresponding foreground in the in-context example is minimal, showcasing its adaptability and strong visual reasoning.

Figure~\ref{fig:visual_examples_med} further showcases examples from two medical imaging datasets, Kvasir and ISIC. In these more complex and domain-specific tasks, E-InMeMo is compared with FMLR (DINOv2). The results reveal that E-InMeMo accurately segments the regions of interest, including gastrointestinal polyps and skin lesions, while FMLR consistently fails to deliver meaningful segmentation outputs. These visual results highlight the capability of E-InMeMo to generalize to medical imagery and reinforce its potential for real-world applications in specialized domains.

\begin{table}[t]
    \centering
    \resizebox{0.75\columnwidth}{!}{
    \begin{tabular}{lccccc}
        \toprule
        Combination & Fold-0 & Fold-1 & Fold-2 & Fold-3 & Mean  \\
        \midrule
        \multicolumn{6}{l}{E-InMeMo variant} \\
        \hspace{5mm}\texttt{I} & 42.28 & 44.86 & 41.10 & 43.24 & 42.87  \\ 
        \hspace{5mm}\texttt{Q} & 40.71 & 43.32 & 39.97 & 40.01 & 41.01  \\ 
        \hspace{5mm}\texttt{I} \& \texttt{Q} & 39.92 & 44.01 & 39.68 & 40.12 & 40.93  \\
        \rowcolor{RoyalBlue!10}\hspace{5mm}\texttt{I} \& \texttt{L} \cellcolor{RoyalBlue!10} (E-InMeMo) & \cellcolor{RoyalBlue!10}\textbf{42.82} & \cellcolor{RoyalBlue!10}\textbf{46.97} & \cellcolor{RoyalBlue!10}\textbf{43.00} & \cellcolor{RoyalBlue!10}\textbf{42.83} & \cellcolor{RoyalBlue!10}\textbf{43.91}  \\
        \bottomrule
    \end{tabular}
    }
    \caption{Segmentation performance for different combinations of images in the canvas to which the learnable prompt is applied. \texttt{I}, \texttt{L}, and \texttt{Q} denote the in-context image, in-context label image, and query image, respectively.}
    \label{tab:location vp}
\end{table}

\textbf{Where should the learnable prompt be applied?} We advocate applying the learnable prompt exclusively to in-context pairs, as they play a crucial role in guiding the model during inference and are central to the effectiveness of visual ICL. To systematically investigate the impact of prompt placement, we three variants of E-InMeMo with different prompt application strategies: (1) applying the prompt only to the in-context image (\texttt{I}); (2) only to the query image (\texttt{Q}); (3) to both the in-context image and query image (\texttt{I} \& \texttt{Q}); and (4) to the in-context image and in-context label (\texttt{I} \& \texttt{L}), which corresponds to the original E-InMeMo formulation.

The evaluation results are reported in Table~\ref{tab:location vp}. Across all variants, the introduction of a learnable prompt improves performance over baseline without the learnable prompt, regardless of where it is applied. Interestingly, while applying the prompt to in-context images alone (\texttt{I}) yields lower performance compared to other E-InMeMo variants, it still surpasses prompt-SelF. This finding reinforces that refining in-context pairs—even minimally—can significantly enhance visual ICL effectiveness.

Surprisingly, the configuration where the prompt is added to both the in-context and query images (\texttt{I} \& \texttt{Q}) performs worse than adding it to either one individually. We attribute this to the input-agnostic nature of the prompt: when the same perturbation is applied to both images, it may hinder the model’s ability to distinguish task-specific cues, thus reducing its capacity to bridge the semantic gap between the query and the support examples.

\begin{figure*}[!t]
  \centering
   \includegraphics[width=1\linewidth]{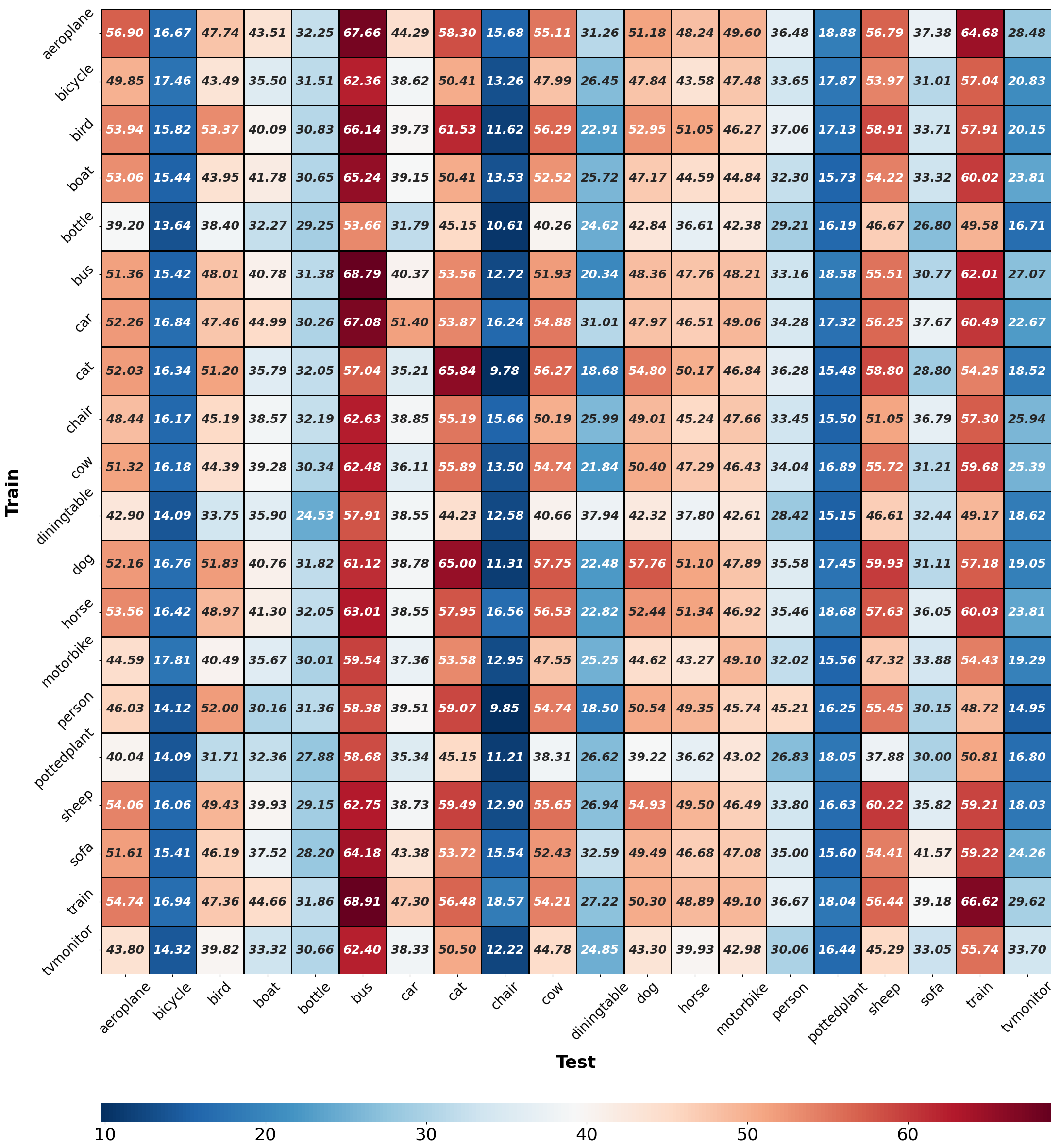}
   \caption{Intra- and inter-class generalization performance in mIoU. The horizontal and vertical axes are the classes used for prediction and training, respectively. The diagonal elements show intra-class performance.}
   \label{fig:cls_base}
\end{figure*}

\textbf{Intra- and inter-class generalizability of E-InMeMo.} We have demonstrated that E-InMeMo performs well on low-quality datasets and now explore its generalizability to unseen classes not included in the training dataset $\mathcal{S}$. To achieve this, we train a learnable prompt for each of the 20 classes. Specifically, let $\mathcal{S}_\omega$ denote the training subset of images and label images in Pascal-5$^i$ for class $\omega$. During E-InMeMo training, an image from $\mathcal{S}_\omega$ is used as a query, paired with an image and label from $\mathcal{S}_\omega$ as an in-context pair. For inter-class generalizability, predictions are made using images in the test set $\mathcal{S}_{\omega'}$ as queries, with in-context pairs retrieved from the training set $\mathcal{S}_\omega$, where $\omega' \neq \omega$. The full results are shown in Table~\ref{tab:main results}. Diagonal values indicate intra-class results. The analysis reveals varying levels of difficulty across classes: (1) Prompts trained on their own class achieve good results for classes such as \texttt{\small{}bus}, \texttt{\small{}cat}, and \texttt{\small{}train}. (2) Prompts trained on their own class perform poorly for classes such as \texttt{\small{}bicycle}, \texttt{\small{}chair}, and \texttt{\small{}pottedplant}.

We also observed that for both challenging and general classes, the overall performance of prompts adapted from different classes aligns with intra-class performance (best viewed vertically in Figure~\ref{fig:cls_base}). For instance, \texttt{\small{}bus}, \texttt{\small{}cat}, \texttt{\small{}sheep}, and \texttt{\small{}train} are the most \textit{generalizable} classes, with prompts trained on different classes achieving high accuracy (IoU close to 50\%). In contrast, classes such as \texttt{\small{}bicycle}, \texttt{\small{}chair}, and \texttt{\small{}pottedplant} are the least generalizable, performing poorly across all prompts. These results suggest that for challenging classes, learned prompts struggle to deliver good performance, whereas for general classes, learned prompts consistently achieve strong results.

In our inter-class analysis (best viewed horizontally in Figure~\ref{fig:cls_base}), we observe that the effectiveness of prompts trained on different classes varies based on the inherent difficulty of each target class. Notably, certain patterns of generalizability emerge across related categories. Within the \texttt{\small{}transportation} super-class—comprising classes such as \texttt{\small{}aeroplane}, \texttt{\small{}train}, and \texttt{\small{}car}—prompts exhibit strong cross-class generalization. For example, prompts trained on \texttt{\small{}aeroplane} generalize well to \texttt{\small{}train} and \texttt{\small{}bus}, and similarly for \texttt{\small{}car}-\texttt{\small{}train} and \texttt{\small{}motorbike}-\texttt{\small{}train} pairs.

In contrast, prompts from \texttt{\small{}transportation} classes tend to generalize poorly to the \texttt{\small{}animal} category (\textit{e.g.}, \texttt{\small{}dog}, \texttt{\small{}horse}). Within the \texttt{\small{}animal} group, most classes show good mutual generalizability, with the exception of \texttt{\small{}cow}, which consistently underperforms. However, prompts from \texttt{\small{}animal} classes also struggle to generalize to \texttt{\small{}transportation} classes.

There are a few interesting exceptions. For instance, the \texttt{\small{}aeroplane} class generalizes surprisingly well to \texttt{\small{}cat}, and \texttt{\small{}bird} performs well when applied to \texttt{\small{}bus}. These anomalies may be attributed to class-specific segmentation properties—for example, \texttt{\small{}cat} and \texttt{\small{}bus} may inherently be easier to segment, regardless of the prompt source.

The mean mIoU score over all pairs of 20 classes is 39.24\%. This score surpasses most methods in Table~\ref{tab:main results} but represents a significant drop compared to E-InMeMo's mean score over all folds. This finding underscores the importance of tuning the learnable prompt for target tasks.

\begin{figure}[!t]
  \centering
   \includegraphics[width=0.7\linewidth]{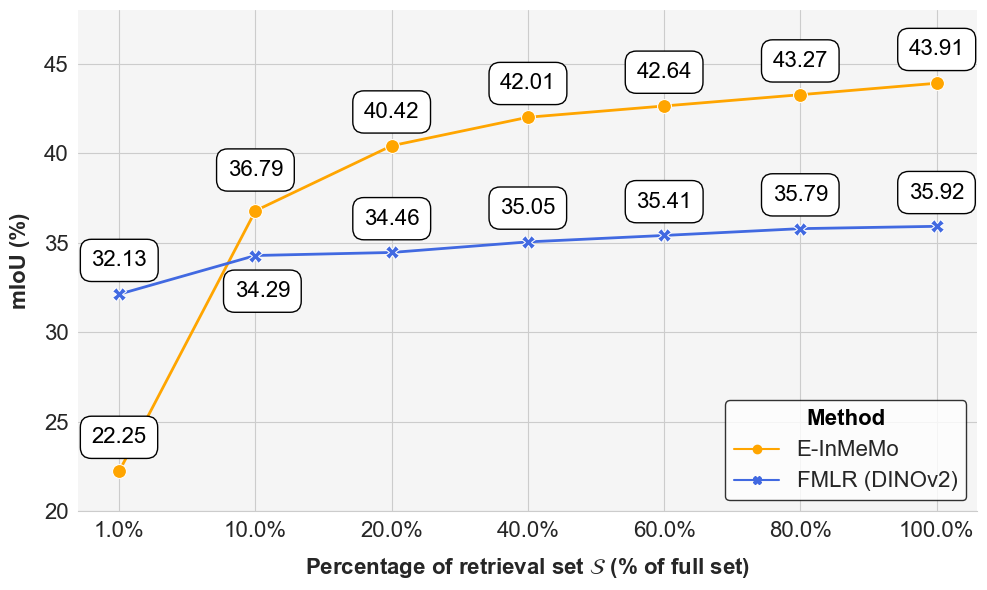}
   \caption{The performance of E-InMeMo and FMLR (DINOv2) in mIoU of each fold for the size of retrieval set $\mathcal{S}$. We annotate the scores in the figure.}
   \label{fig:retrieval_set}
\end{figure}

\textbf{Is E-InMeMo sensitive to the size of retrieval dataset?} Following \cite{supicl}, we wonder if E-InMeMo is sensitive to the size of the retrieval set $\mathcal{S}$ or not. So we use 7 different subsets (1\%, 10\%, 20\%, 40\%, 60\%, 80\% and 100\%) on each fold as retrieval sets to be used for retrieval and as training sets for E-InMeMo training. The results are shown in Figure~\ref{fig:retrieval_set}. It is obvious that both FMLR (DINOv2) and E-InMeMo benefit from larger retrieval set sizes. There is a clear trend in the growth of E-InMeMo, especially when going from 1.0\% to 10.0\%, there is a significant increase in mIoU. E-InMeMo is more sensitive to the data, and when the retrieval set is increased from 1\% to 100\%, its mIoU improves by 21.66 points. FMLR (DINOv2), on the other hand, is not as sensitive to the size of the retrieval set and improves by only 3.79 points. It should be noted that E-InMeMo needs only 10\% of the retrieval set to achieve significantly better results than FMLR (DINOv2) at 100\% retrieval set with parameter-efficient fine-tuning, which side by side emphasizes that E-InMeMo is a data efficient method.

\begin{figure}[!t]
  \centering
   \includegraphics[width=0.7\linewidth]{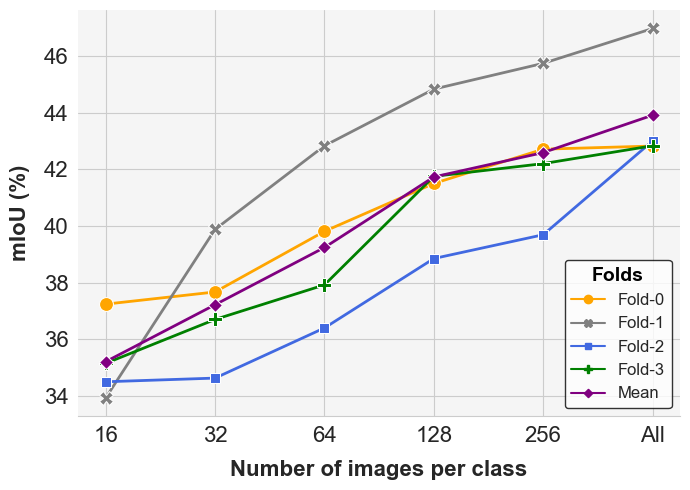}
   \caption{The performance of E-InMeMo in mIoU of each fold for the number of images per class in $\mathcal{S}$. All means to use all images in the training set.}
   \label{fig:fsl_result}
\end{figure}

\textbf{Is E-InMeMo sensitive to dataset size?} To better understand how the performance of E-InMeMo scales with the size of the dataset $\mathcal{S}$, we investigate its sensitivity to varying data volumes. Given the lightweight and efficient nature of the learnable prompt, we systematically varied the dataset size by randomly selecting 16, 32, 64, 128, and 256 images per class for each fold. The performance trends are illustrated in Figure~\ref{fig:fsl_result}.

The experimental results indicate that E-InMeMo outperforms the baseline mIoU score of 35.92\% once the dataset includes at least 64 images per class, achieving 37.22\% at that point. Across all folds, performance generally improves as more data is provided. For instance, in Fold-0, E-InMeMo reaches 37.24\% with only 16 images per class and shows rapid gains starting from 32 images, eventually saturating at 256. Both Fold-1 and Fold-2 demonstrate steady and significant improvements with increasing data volume.

In Fold-3, a more complex fold, a notable jump in performance is observed between 64 and 128 images per class. Beyond this, the improvement plateaus, with only marginal gains when utilizing the full dataset.

Overall, these results suggest that E-InMeMo requires fewer training examples for simpler folds, yet benefits from larger datasets in more complex scenarios. This data-efficiency makes E-InMeMo particularly suitable for tasks where labeled data is limited, while still being capable of leveraging larger datasets when available to further boost performance.

\begin{table}[t]
    \centering
    \resizebox{0.8\columnwidth}{!}{
    \begin{tabular}{lcccccc}
        \toprule
        Padding size & Para. & Fold-0 & Fold-1 & Fold-2 & Fold-3 & Mean  \\ 
        \midrule
        \hspace{5mm} 5  & 9,780 & 41.31 & 47.12 & 41.37 & 41.14 & 42.73  \\
        \hspace{5mm} 10 & 18,960 & \textbf{43.59} & \textbf{47.86} & 41.46 & 42.65 & 43.89  \\
        \rowcolor{RoyalBlue!10}\hspace{5mm} 15 (E-InMeMo) & 27,540 & 42.82 & 46.97 & \textbf{43.00} & \textbf{42.83} & \textbf{43.91}  \\
        \hspace{5mm} 20 & 35,520 & 39.23 & 42.81 & 37.77 & 41.06 & 40.22  \\ 
        \hspace{5mm} 25 & 42,900 & 42.22 & 45.96 & 37.96 & 39.72 & 41.47  \\ 
        \hspace{5mm} 30 & 49,680 & 31.53 & 40.62 & 36.10 & 39.26 & 36.88  \\
        \bottomrule
    \end{tabular}
    }
    \caption{The mIoU scores for varying padding sizes of prompt enhancer $t_\phi$. The Para. represents the number of tunable parameters. The highest score in each fold is marked in \textbf{bold}.}
    \label{tab:padding size}
\end{table}

\textbf{Effect of Padding Size in the Prompt Enhancer $t_\phi$}
In our implementation, the padding size for E-InMeMo’s learnable prompt enhancer $t_\phi$ is set to 15 pixels. To investigate the influence of different padding sizes on model performance, we conducted a comprehensive evaluation, with the results summarized in Table~\ref{tab:padding size}.

E-InMeMo consistently outperformed all baseline methods across different padding sizes in terms of Mean performance. As the padding size increased from 5 to 15, we observed a steady improvement, suggesting that a moderate spatial footprint for the prompt allows the model to better guide the in-context pair without excessive interference.

However, further increasing the padding beyond 15 pixels led to diminishing returns. Specifically, increasing the padding size to 20 resulted in a noticeable drop in performance—from 43.91\% to 40.22\% on Mean. Although a small gain was observed when moving from 20 to 25 pixels, performance remained below the optimal. When the padding was increased to 30, the performance continued to decline.

Interestingly, at a padding size of 10, the model achieved the best results on Fold-0 (43.59\%) and Fold-1 (47.86\%), yet overall Mean performance was slightly lower than that of padding size 15. This suggests that while smaller prompts can be beneficial in specific scenarios, they may lack the generalization capacity required across more diverse or complex folds. We hypothesize that larger padding sizes (\textit{e.g.}, larger than 20) cause the learnable prompt to occupy too much of the original image, potentially overwriting critical visual information from the in-context pair. This over-perturbation compromises the prompt's guiding ability and leads to a drop in performance.

In summary, our findings indicate that a padding size of 15 offers the best trade-off between expressive capacity and preservation of the original visual content, making it the empirically optimal setting for E-InMeMo. Expanding the padding further introduces unnecessary parameters and can degrade performance.



\section{Discussion}
We proposed E-InMeMo, a parameter-efficient method for applying vision foundation models to specific tasks and domains. E-InMeMo has broad applicability in fields such as medicine and environmental science, where ease of adaptation and high accuracy are essential. The basic performance of visual in-context learning (ICL) can be further enhanced depending on the strength of the underlying vision foundation models \cite{maevqgan, supicl}. Leveraging powerful models and integrating E-InMeMo with data- and parameter-efficient strategies could be a promising direction for real-world deployment.

\textbf{Limitations.} Despite its strengths, E-InMeMo requires a minimum of 32 images per class to achieve competitive performance relative to the baseline. Furthermore, the learned prompt is task-specific, a prompt trained for one class or task does not generalize well to others. Therefore, designing dedicated prompts for each target task remains essential to maximizing performance.

\section{Conclusion}

E-InMeMo achieves SOTA performance on two downstream tasks by introducing a learnable prompt to the in-context pairs, a lightweight yet powerful approach that enhances the effectiveness of visual in-context learning (ICL). This learnable prompt improves the model’s ability to recover fine-grained details in predictions and mitigates the negative impact of low-quality or semantically distant in-context pairs. Furthermore, our experiments demonstrate that E-InMeMo exhibits robustness to domain shifts, such as transferring from the Pascal dataset to COCO.

\section*{Acknowledgements}
This work was supported by JSPS KAKENHI Grant No. JP23H00497, JST CREST Grant No. JPMJCR20D3, JST ACT-X Grant No. JPMJAX24C8, JSPS KAKENHI No. 24K20795, and FOREST Grant No. JPMJFR216O.

\bibliographystyle{elsarticle-num}
\bibliography{main}

\end{document}